# EXPERIMENTAL CHARACTERIZATION OF ROBOT ARM RIGIDITY IN ORDER TO BE USED IN MACHINING OPERATION

Jean-Yves K'NEVEZ[1], Mehdi CHERIF[2], Miron ZAPCIU[3], Alain GERARD[4]

***Abstract:*** *Attempts to install a rotating tool at the end of a robot arm poly-articulated date back twenty years, but these robots were not designed for that. Indeed, two essential features are necessary for machining: high rigidity and precision in a given workspace. The experimental results presented are the dynamic identification of a poly-articulated robot equipped with an integrated spindle. This study aims to highlight the influence of the geometric configuration of the robot arm on the overall stiffness of the system. The spindle is taken into account as an additional weight on board but also as a dynamical excitation for the robot KUKA KR_240_2. Study of the robotic machining vibrations shows the suitable directions of movement in milling process.*

## 1. INTRODUCTION

Industrial robots are changing the face of milling operation. Up to this point, milling has been accomplished with special milling and CNC machines. The robots offer the following advantages over these traditional milling methods:

- *Flexibility* - 6-axis for typical articulated robot offers more movement flexibility than a normal milling machine. A robot can mill a complex surface of the part.

- *Throughput* - Milling with a robotic arm can increase overall throughput. A robot is more consistent and accurate. With fewer mistakes, and less time spent repositioning a robotic arm mills faster.

- *Right Touch* - Many of the materials used for prototyping and molds are soft - clay, foam, REN board. A robotic arm is well-suited for responding to and working with all types of mediums.

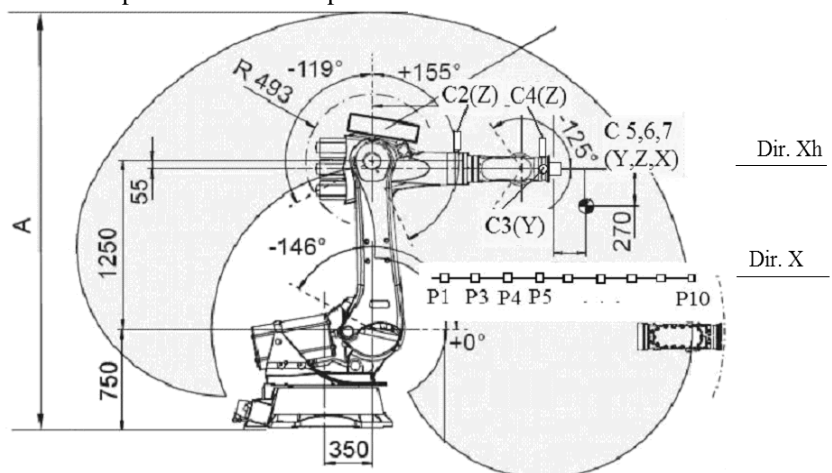
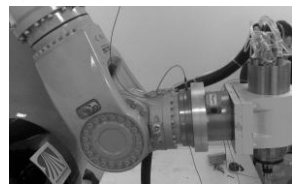

**Fig.1** General view and main characteristics of the robot arm Kuka KR 240-2.

| Type | KR 240-2 |
|---|---|
| Payload | 240 kg |
| Supplementary load | |
| Max. supplementary load | |
| Total distributed load | 640 kg |
| Weight (excl. controller) approx. | 1267 kg |

[1] University of Bordeaux, LMP (Bordeaux 1- CNRS UMR 5469), 351, Cours de la Libération, 33405 Talence, France
Tel. + (33) (0)5 40 00 62 22 , Fax.:+(33) (0)5 40 00 69 64 ; E-mail address: jean-yves.knevez@u-bordeaux1.fr
[2] University of Bordeaux , LGM²B – IUT Bordeaux 1, 15 rue Naudet, 33175 Gradignan Cedex France
[3] University Politehnica of Bucharest, MSP department, Spl Independentei 313, 060032, Romania
Tel./Fax. +40 21 402 9724; E-mail address: zapcium@yahoo.com
[4] University of Bordeaux, LMP (Bordeaux 1- CNRS UMR 5469), 351, Cours de la Libération, 33405 Talence, France

- *Affordability* - Milling with a robotic arm is extremely economical. Unlike milling machines, robots can be reassigned to perform other assignments in a shop - arc welding, material handling, etc. They can improve throughput, saving valuable time and increasing production. In addition, a robotic arm can handle more of the milling task without needing human intervention.

Many producers, for example Kuka (Fig.1) offer application-specific components and tools for deployment of a robot as a machine tool for milling tasks [1].

Vibration of arm robot structure is the major limitation of robotic machining capacities. The presence of the low frequency modes will shake the entire robot body and cause instability of the dynamic system during machining.

The stiffness of the CNC machine is usually hundreds of times larger than process stiffness and mode coupling chatter rarely happen. For robot, the difference is only 5–10 times. This mode coupling effect is the dominant reason for structure vibration in robotic machining process [2].

The relative orientation of the force vector and the principle stiffness axes are the dominant factors that affect the stability of machining process using robots. Methods such as changing the feed direction, using different robot configuration or changing another type of tool are all worth trying. Based on the practical investigations, this research leads to a deeper understanding of the unstable phenomenon in robotic machining process and provides a guideline as well as practical solutions to avoid such problems.

## 2. CHARACTERIZATION OF THE STIFFNESS OF KUKA KR-240-2 ROBOT ARM

Milling process is specially designed for machining tasks using an electrically-driven spindle. It is used particularly with lightweight materials such as plastic, composite or rigid foamed material. From the HSC spindle and its controller to the special milling software, is possible to quick and easy setup the robot as a powerful milling unit [3].

Experimental research involves the application of three unidirectional accelerometers C2, C3 and C4, on the robot arm (see the figure 1) and a three-axial sensor C5 (Y), C6 (Z) and C7 (X) on the mechanical interface of the robot. Responses of the robot arm were made considering the transfer function using hammer impact method. The signal acquired was the apparent mass in dependence with frequency. Were used 10 measurement points P1 ... P10 distance between these points being equal and having a value of 200 mm.

Details of the signals acquired with C4 sensor in the range 0-200 Hz are detailed in the figure 2. Robot arm was in X1 position (P1) and two situations were experimented: arm with brake or without brake in function. Significant frequencies were 23 Hz, 80 Hz and 95 Hz and it was found that there are no differences between the two operating situations (brake has no influence on the robot arm rigidity).

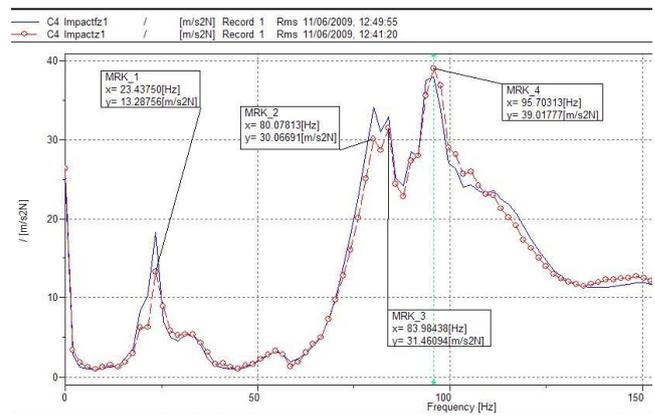

**Fig.2.** Signals acquired with C4 sensor in the range 0-200 Hz; brake / without brake in function

Stiffness after X axis is the least important in the milling process (the deformation is greatest on the plan ZOY). Figure 3 represents the X axis sensitivity to the impact by Y, considering measurements along the X direction, in the points P1 ... 10. Largest mechanical coupling between X and Y axes is found in section P1.

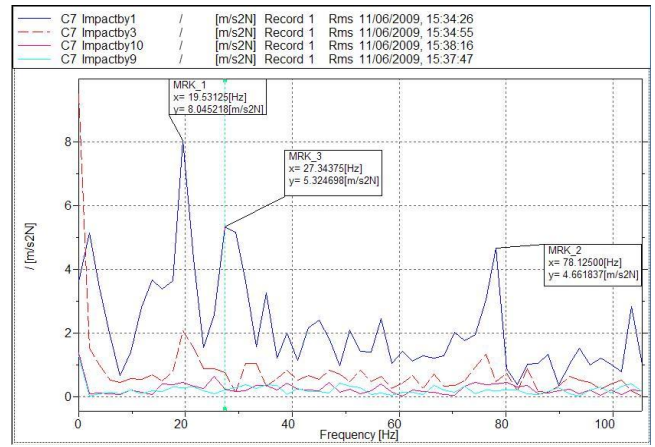

**Fig.3.** The coupling between the X and Y axes in positions P1, … , P10

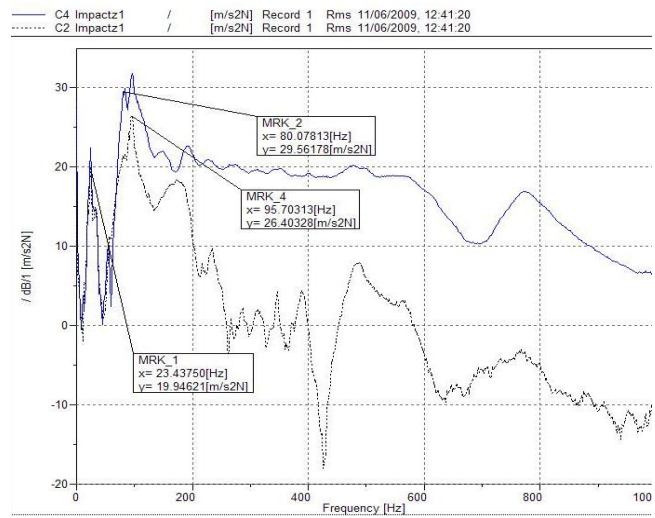

**Fig.4.** Comparison between the signal of C2 and C4, in Z direction

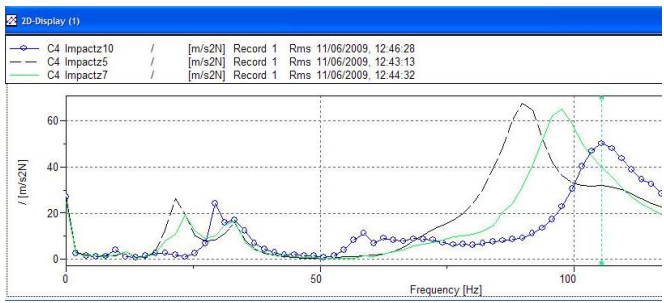

**Fig. 5.** Details on the frequency shift (stiffness variation) on the Z direction (comparison of the impact signals acquired in positions 5, 7 and 10)

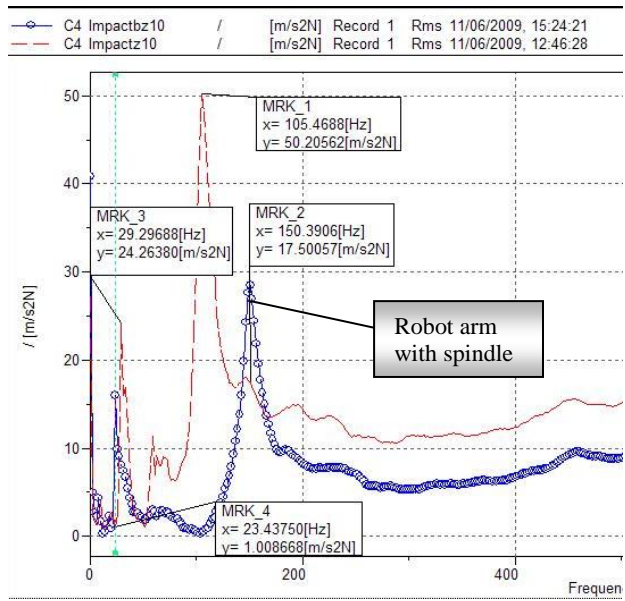

**Fig. 6.** Comparison of the values of first two natural frequencies of the robot arm (with & without spindle); position 10 - direction Z.

The comparison between the signal of C2 and C4, in Z direction, for the robot without spindle is presented in the figure 4. The robot arm segment situated near the base of the robot is more sensitive to low frequencies.

Another interesting experimental research was to check the robot arm stiffness when the arm is extended (between the points P1 ... P10).

Figure 5 shows that the signal is acquired the same shape (same number of degrees of freedom) but the stiffness and apparent mass change. Structural natural frequency increases as the arm extends to the farthest point P10.

Spindle mass influences the dynamic behavior of the robot arm. To study this influence (Fig. 6) were made two determinations of the transfer function in position 10 (elongated arm). The first frequency decreases caused by the mass of the spindles that was important -53 kg.

Because the first natural frequency of the robot arm has relatively little value, was studied in detail the range of these values.

Figure 7 details that when the robot arm extends from point P1 to point P10, the values of the first frequencies in direction Y are in the field of 15,6-19,5 Hz (higher values correspond to the extended arm).

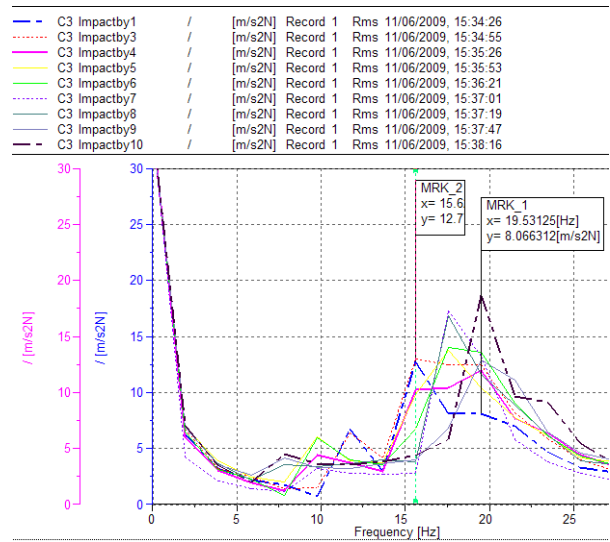

**Fig. 7.** Comparison of the values of first natural frequency of the robot arm with integrated spindle fixed (direction Y; 15,6 Hz (P1) …. 19,5 Hz (P10))

Considering Z direction, the favorable field of frequency was 175 Hz ...1250 Hz. The same types of measurements were performed in Y direction, being obtained a favorable area for use in the range 175 ... 1750 Hz.

Dynamic behavior of the spindle in the domain 0 - 5000 Hz is presented using FFT spectrum in the Figure 8. The frequency of 498 Hz is the frequency of rotation of the spindle. First natural frequency of the spindle (robot arm attached) is about 2000 Hz.

Using the speed of the spindle at 5000 rpm, the level of vibration in the vertical plane (Z direction) is 50% of the level in horizontally plane - Y direction (Fig. 9); sensors were placed on the terminal of the robot arm.

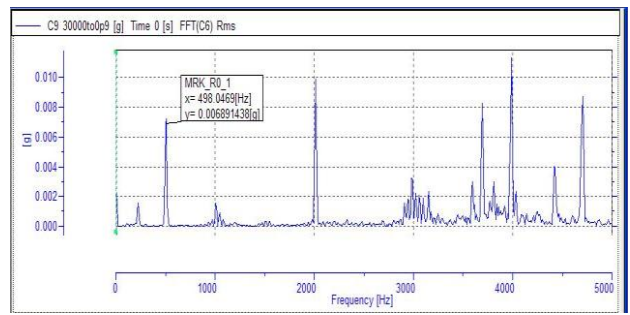

**Fig. 8.** FFT of the spindle in the range 0…5000 Hz

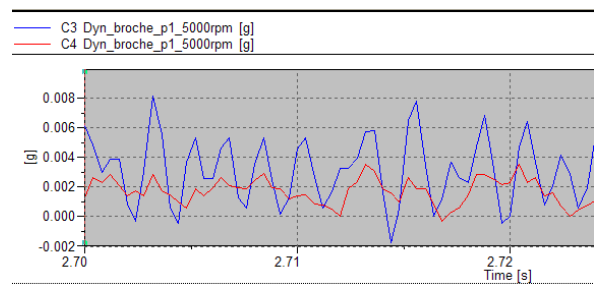

**Fig.9.** Level of vibrations obtained with sensors C3 (Y) and C4 (Z).

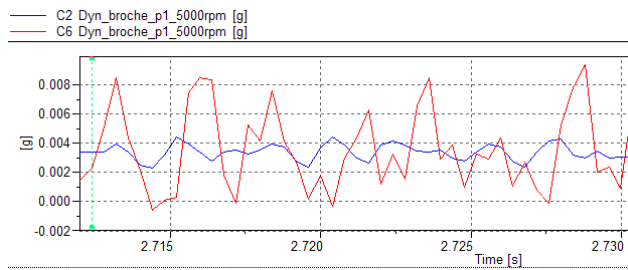

**Fig.10.** Level of vibrations in the aria of C2 and C6 sensors (Z direction).

At the same speed, 5000 rpm, in the vertical plane (Z direction), the level of vibration on the sensor area C2 (the robot arm 2) is 25% of the level of vibration obtained with the sensor C6 (Fig. 10).

## 3. ROBOT ARM RIGIDITY ALONG TWO PERPENDICULAR DIRECTIONS

Another objective of this work was to determine the stiffness of the robot arm along two perpendicular directions. XH direction corresponds of X axis to an altitude of 1250 mm from the robot base and 2000 mm above the ground. Other direction, Y7, corresponds of Y axis and intersects XH in the point P7. The comparison of the stiffness of the robot arm in different positions along the directions XH and Y7 are presented in the figures 11 and 12.

The results were obtained based on the relation (1), considering the apparent mass corresponding to their first natural frequency (ideally [4]).

$$f^2 \text{ const.} = k_{equiv} / m_{equiv} \quad (1)$$

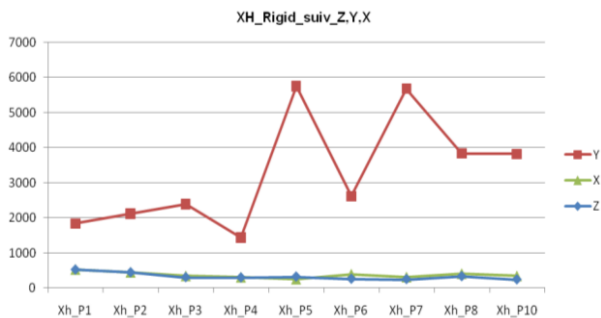

**Fig. 11.** Stiffness comparison between the different positions along the XH direction

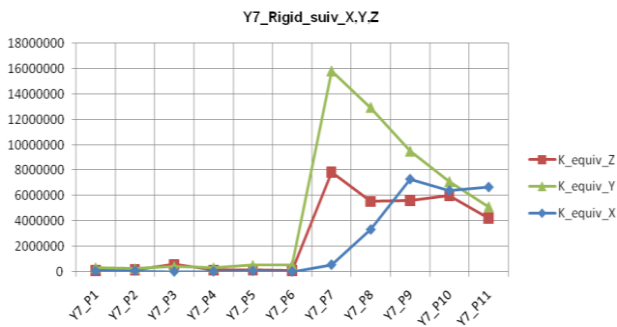

**Fig. 12.** Stiffness comparison between the different positions along the Y7 direction

It noted that after the Y direction, the robot arm stiffness varies greatly between points P6 ... P10. A map of the stiffness of the specific robot arm structure is needed to determine the most convenient position for the work-piece within the workspace of the robot.

## 4. CONCLUSIONS

The speed domain of the milling spindle is selected and influenced by the position of the robot arm due to the different stiffness in the area of work [5].

Using 5000 rpm, the vibration level (amplitude of acceleration) in the vertical plane (Z direction) is 50% of the horizontal amplitude (Y direction) ; in the same plane, the vibration level in the sensor area C2 (the robot arm 2) is 25% of the vibration level of the fixation point of the spindle (point P1). The range of the recommended frequencies in the entire working area is: 175 Hz-700 Hz (10.500 – 42.000 rpm).

For more rigidity it is possible to use the spindle in the horizontal position (axis of the tool along the Y axis of the robot). The stiffness along the X axis is 500% of the stiffness along Y and Z, between points P1 ... P4, and much more rigid between P5 and P10 (arm elongated).

When the robot moves along the X axis, the stiffness of the robot arm has a tendency to decrease (in the vertical and horizontal planes) and axial stiffness increases between points P5 … P10. Following Y axis, the stiffness increases when the robot moves the milling spindle between points P7 ... P11 (Y positive). Following the direction X-X, the first natural frequency of the robot (with spindle in vertical position) is inside the range 17 Hz - 23 Hz.

A novelty in this paper is to analyze the variability of the robot arm stiffness in order to determine areas where stiffness has values with large variations. The work was done as a result of collaboration by a team of specialists from the Machines and Production Systems lab from Bucharest and the laboratories LGM$^2$B and LMP from University of Bordeaux 1.